\documentclass[conference]{IEEEtran}
\IEEEoverridecommandlockouts
\usepackage{cite}
\usepackage{amsmath,amssymb,amsfonts}
\usepackage{algorithmic}
\usepackage{graphicx}
\usepackage{textcomp}
\usepackage{xcolor}
\usepackage{booktabs}
\usepackage{multirow}

\def\BibTeX{{\rm B\kern-.05em{\sc i\kern-.025em b}\kern-.08em
    T\kern-.1667em\lower.7ex\hbox{E}\kern-.125emX}}
\begin{document}

\title{Dynamic Visual-semantic Alignment for Zero-shot Learning with Ambiguous Labels
\thanks{$^{\ast}$Corresponding author (fan\_jinfu@163.com). This work was supported by the National Natural Science Foundation of China (Nos. 62401309, 62401362, 62576183), the China Postdoctoral Science Foundations (No. 2025M771692), the Natural Science Foundation of Shandong Province and Qingdao (No. 2R2024QF119, No. 24-4-4-zrjj-89-jch), and open project (No. 2024PY030).}
}


\author{
	\IEEEauthorblockN{
		Jiangnan Li$^{1}$, 
		Linqing Huang$^{2}$, 
        Xiaowen Yan$^{1}$,
		Min Gan$^{1}$,
        Wenpeng Lu$^{3}$,
        Jinfu Fan$^{1,\ast}$
    }    
     
  \IEEEauthorblockA{%
  \begin{tabular}{c}
    $^{1}$ College of Computer Science and Technology, Qingdao University \\
    $^{2}$ School of Computer Science, Shanghai JiaoTong University \\
    $^{3}$ Key Laboratory of Computing Power Network and Information Security, Ministry of Education, \\Shandong Computer Science Center, Qilu University of Technology (Shandong Academy of Sciences) 
    \vspace{-0.3cm}
  \end{tabular}%
}
}

\maketitle

\begin{abstract}

Zero-shot learning (ZSL) aims to recognize unseen classes without visual instances. However, existing methods usually assume clean labels, overlooking real-world label noise and ambiguity, which degrades performance. To bridge this gap, we propose the Dynamic Visual-semantic Alignment (DVSA), a robust ZSL framework for learning from ambiguous labels. DVSA uses a bidirectional visual–semantic alignment module with attention to mutually calibrate visual features and attribute prototypes, and a contrastive optimization grounded in Mutual Information (MI) at the attribute level to strengthen discriminative, semantically consistent attributes. In addition, a dynamic label disambiguation mechanism iteratively corrects noisy supervision while preserving semantic consistency, narrowing the instance-label gap, and improving generalization. Extensive experiments on standard benchmarks verify that DVSA achieves stronger performance under ambiguous supervision.
\end{abstract}


\section{Introduction}
\label{sec:intro}

\emph{Zero-shot learning} (ZSL) has emerged as an important paradigm for visual recognition that aims to mimic human cognition by recognizing unseen classes without requiring labeled visual instances \cite{lampert2009learning}. By leveraging semantic side information shared across classes such as attributes \cite{lampert2009learning}, word embeddings \cite{socher2013zero}, and textual descriptions \cite{reed2016learning}, ZSL transfers knowledge from seen to unseen classes and thereby alleviates the need for exhaustive manual annotation.

\begin{figure}
  \centering
  \includegraphics[width=.95\columnwidth]{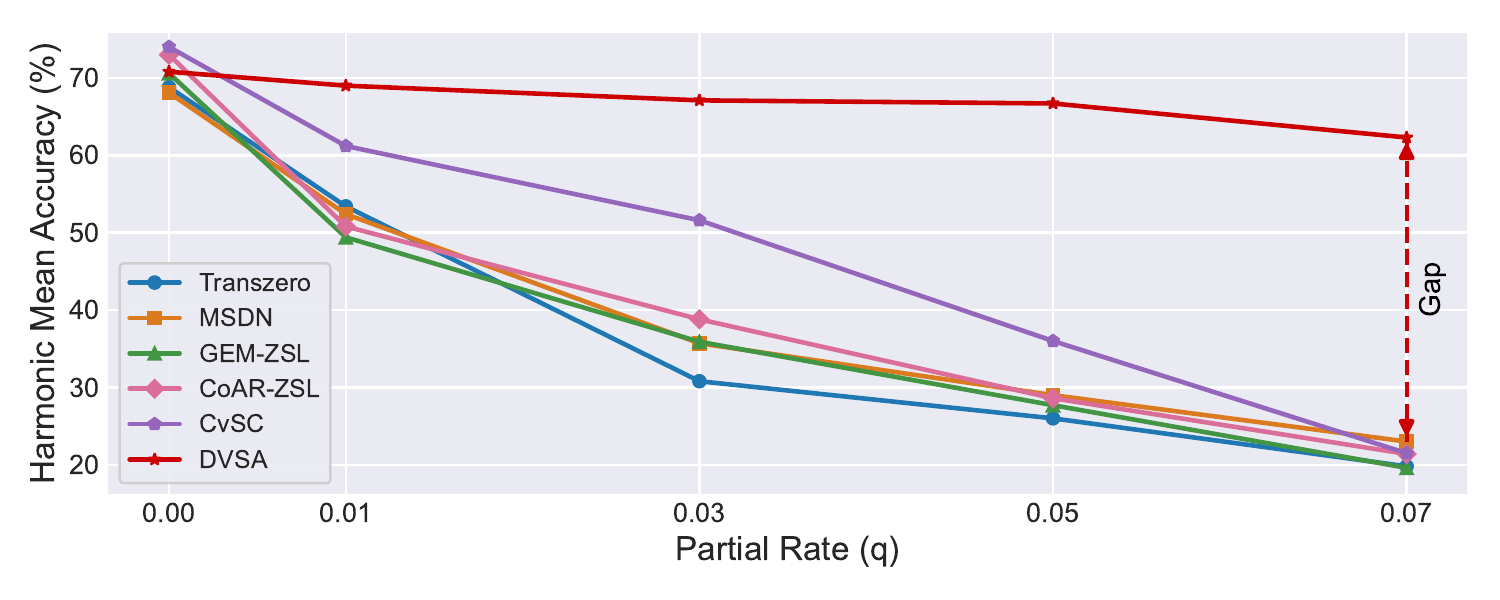}
  \vskip -0.1in
  \caption{How Ambiguous Labels Affect ZSL Performance.}
  \label{fig:intro}
  \vskip -0.15in
\end{figure}

Existing approaches can be broadly classified into generative and embedding-based methods. Generative methods use auxiliary generators \cite{li2019leveraging,schonfeld2019generalized,ye2025zerodiff} to synthesize visual features for unseen classes. Although effective, their reliance on additional external data transforms ZSL into a fully supervised task. In contrast, embedding-based methods instead learn a shared space where visual features and semantic prototypes are aligned for nearest-neighbor prediction \cite{frome2013devise,romera2015embarrassingly,liu2021goal}. Despite its success, a predominant limitation of existing methods is their reliance on the strong assumption that the training instances from seen classes are annotated with accurate and reliable class labels. This assumption is often unrealistic in practice, especially for fine-grained species recognition or large-scale long-tail scenarios, where annotations are more prone to ambiguity and noise. Similar robustness challenges have also been discussed in incomplete-data classification settings \cite{huang2026incomplete}. As illustrated in Figure~\ref{fig:intro}, when we inject such ambiguity into the CUB dataset, state-of-the-art ZSL models suffer a sharp performance drop, \textbf{\emph{revealing a clear vulnerability of current ZSL pipelines to imperfect supervision.}}

To alleviate the annotation burden, weakly supervised learning methods such as \emph{Partial label learning} (PLL) \cite{zhang2015solving,fan2024kmt} have been explored. PLL allows training instances to be annotated with ambiguous labels, where each instance may have multiple candidate labels, with only one being correct. By exploiting relationships among candidate labels of similar instances, PLL alleviates the detrimental effects of label ambiguity on classifier performance.

Building on these insights, we aim to combine the strengths of ZSL and PLL to mitigate ambiguous labels in training data while effectively predicting unseen classes. To this end, we propose the \emph{Dynamic Visual-semantic Alignment} (DVSA) framework for zero-shot learning with ambiguous labels. DVSA first aligns visual features and attribute vectors via a bidirectional attention module, ensuring mutual reinforcement between visual and semantic representations. Visual-to-Attribute (VTA) block refines attribute prototypes using visual cues, while an Attribute-to-Visual (ATV) block calibrates visual features conditioned on the updated attributes. An attribute-level mutual information (MI) based contrastive objective encourages the selected attributes to maximally preserve semantic consistency while increasing separation from negatives, yielding more discriminative and noise-resilient attribute representations. Finally, a dynamic label disambiguation mechanism iteratively updates soft label distributions over candidate labels using both visual and semantic predictions, and feeds the refined supervision back into training, narrowing the semantic gap between instances and candidate labels.

\begin{figure*}
  \centering
  \includegraphics[width=.85\textwidth]{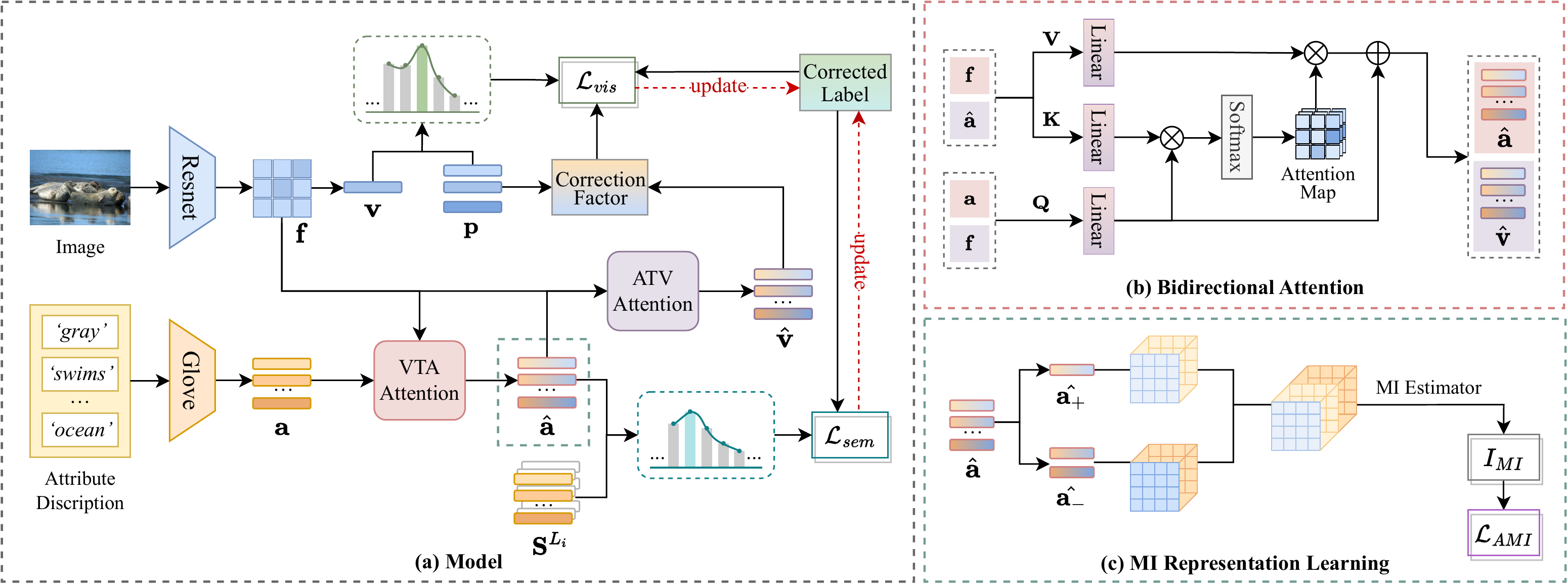}
  \caption{The framework of our proposed DVSA. DVSA processes attributes and images with pre-trained encoders and calibrates features via a bidirectional attention module. As shown in (b), the VTA block selectively enhances attribute vectors $\mathbf{a}$ under the guidance of visual features $\mathbf{f}$, while the ATV block refines visual features $\mathbf{f}$ conditioned on the updated attributes $\hat{\mathbf{a}}$. The MI-based optimization in (c) further strengthens the semantic correlations encoded in $\hat{\mathbf{a}}$.}
  \label{fig:model}
  \vskip -0.1in
\end{figure*}

In summary, our main contributions are threefold:
\begin{itemize}
  \item We devise a novel visual-semantic alignment module that combines bidirectional attention and MI optimization, enabling more discriminative and robust cross-modal alignment under ambiguous supervision.
  \item We propose a novel dynamic label disambiguation mechanism to progressively refine supervision signals, which requires no clean label assumptions.
  \item Extensive experiments demonstrate the effectiveness of our DVSA with superior robustness to label ambiguity.
\end{itemize}

\section{Related Work}

\textbf{\emph{Zero-shot learning}} (ZSL) transfers semantic knowledge from seen to unseen classes via shared representations. Generative-based methods synthesize unseen features with auxiliary generators (e.g., GAN \cite{li2019leveraging}, VAE \cite{schonfeld2019generalized}, diffusion \cite{ye2025zerodiff}). While embedding-based methods learn a joint space aligning visual features and semantic prototypes for nearest-neighbor. Although recent models leverage attention \cite{liu2021goal}, graphs \cite{liu2020attribute}, and transformers \cite{chen2022transzero} to obtain more discriminative features, they still degrade notably under label noise or ambiguity.

\textbf{\emph{Partial label learning}} (PLL) associates each instance with a candidate label set, of which only one is correct. Conventional PLL disambiguates by exploiting similarity among instances and candidate labels through averaging \cite{zhang2015solving} or identification \cite{zhou2016partial} strategies. Deep PLL further couples representation learning with disambiguation using regularization \cite{wang2022pico+}, data augmentation \cite{chai2019large}, and dynamic loss reweighting \cite{wen2021leveraged}, substantially improving robustness to noisy supervision.

\textbf{\emph{Mutual information}} (MI) measures statistical dependency and has become a key tool for representation learning. MINE \cite{belghazi2018mine} introduced neural estimation of MI for high-dimensional continuous variables. Deep InfoMax \cite{hjelm2018learning} demonstrates that maximizing MI between local and global representations improves downstream task performance. In ZSL and PLL, MI-based objectives have been used to strengthen alignment and guide disambiguation \cite{fan2023graphdpi}, demonstrating the effectiveness and promising prospects.

\section{Method}
\label{method}

\noindent\textbf{{Problem Setting.}}
Let the instance space be $\mathcal{X}=\mathcal{X}^s \cup \mathcal{X}^u$. The seen label space is $\mathcal{Y}^{s}=\{y_{c}\}_{c=1}^{Q}$, where each class $y_{c}$ has an semantic vector $\mathbf{S}^{c}=[s_{1}^{c},\ldots,s_{K}^{c}]^{\top}$ with $K$ attributes. The training set $\mathcal{D}=\left\{ (\mathbf{x}_{i},  L_{i}, \mathbf{S}^{L_{i}}) |1\leq i\leq N \right\}$ contains $N$ instances, $L_{i}\subseteq\mathcal{Y}^{s}$ is a candidate label set for $\mathbf{x}_i$ with cardinality $|L_{i}|$, and $\mathbf{S}^{L_{i}}$ collects the attribute vectors of labels in $L_{i}$. $L_i$ can also be represented by a label vector $\mathbf{l}_ i$, where $l_{ic} \in\{0,1\}$ denotes whether the label $y_{c}$ is present in the candidate labels (`1') or not (`0'). During training, the true label of an instance $\mathbf{x}_{i}$ is only partially observed through its candidate label set $L_{i}$, and the model must leverage $\mathbf{S}^{L_{i}}$ to learn from such ambiguous supervision. 
In the conventional ZSL (CZSL), the task is to recognize only unseen classes, whereas in generalized ZSL (GZSL) instances may come from both seen and unseen classes, with $\mathcal{Y}^{s} \cap \mathcal{Y}^{u}=\emptyset$. The objective of our model is to learn a classifier $\boldsymbol{f}: \mathcal{X} \mapsto \mathcal{Y}^{s} \cup \mathcal{Y}^{u}$ that can effectively remove the influence of noise in the candidate labels, accurately predicting the true labels for instances.

\subsection{Visual-semantic Alignment Module}

\subsubsection{Bidirectional Attention}

\paragraph{Visual-to-Attribute Attention Block}
The VTA block refines attribute vectors using visual features extracted from the image. As shown in the pink rectangular box in Fig.~\ref{fig:model}(b), original attribute representation $\mathbf{a}\in\mathbb{R}^{K\times d_{w2v}}$ act as the query and perform attention to regional visual features $\mathbf{f}\in\mathbb{R}^{D\times d_{v}}$ from ResNet101. We obtain query, key, and value projections by linear transformation layers,
\begin{equation} 
    \mathbf{Q}_1=\mathbf{a} \mathbf{W}_{{Q}_1}, \quad 
    \mathbf{K}_1=\mathbf{f} \mathbf{W}_{{K}_1}, \quad
    \mathbf{V}_1=\mathbf{f} \mathbf{W}_{{V}_1},
\end{equation} 
where $\mathbf{W}_{{Q}_1}\in\mathbb{R}^{d_{w2v}\times d},\mathbf{W}_{{K}_1}\in\mathbb{R}^{d_v\times d}$ and $\mathbf{W}_{{V}_1}\in\mathbb{R}^{d_v\times d}$ are learnable parameter matrices. To capture the relationships between the attributes and the visual features, the correlation between the $\mathbf{Q}_1$ and $\mathbf{K}_1$ is computed via the dot product, followed by a SoftMax operation to produce an attention map. Then, the attention map is applied to $\mathbf{V}_1$ with a residual connection to form instance-specific attribute prototypes,
\begin{equation}
    \hat{\mathbf{a}}_m= \text{Softmax}\left(\frac{\mathbf{Q}_1 \times \mathbf{K}_1^\top}{\sqrt{d}}\right) \mathbf{V}_1+\mathbf{Q}_1.
\end{equation} 
Finally, we project back to the original attribute dimension with $\mathbf{W}_{O}\in\mathbb{R}^{d\times d_v}$ to obtain refined attributes representations,
\begin{equation}
    \hat{\mathbf{a}}=\hat{\mathbf{a}}_m\mathbf{W}_{O}.
\end{equation} 

By incorporating the VTA block, DVSA is capable of progressively refining the visual-semantic alignment and extracting attribute representations related to instances during the iterative process of focusing on visual-semantic correlations. This module narrows the semantic gap between visual features and attribute prototypes by using discriminative visual cues to guide the calibration of attribute vectors.

\paragraph{Attribute-to-Visual Attention Block}
The ATV block refines visual features based on the updated attribute representations obtained from the VTA block. This block operates in the reverse direction compared to the VTA attention block. As shown in the purple rectangular box in Figure~\ref{fig:model} (b), the visual features $\mathbf{f}$ now act as the query, while the updated attribute representations $\hat{\mathbf{a}}$ serve as both the key and value. Similarly, unambiguous and transferable visual representations is
\begin{equation} 
    \hat{\mathbf{v}}= \left(\text{Softmax}\left(\frac{\mathbf{f} \mathbf{W}_{{Q}_2} \times (\hat{\mathbf{a}} \mathbf{W}_{{K}_2})^\top}{\sqrt{d}}\right) \hat{\mathbf{a}} \mathbf{W}_{{V}_2}+\mathbf{f} \mathbf{W}_{{Q}_2}\right)\mathbf{W}_{O}.
\end{equation} 

Together with VTA, ATV forms a bidirectional alignment loop. VTA distills attribute representations from visual evidence, while ATV projects these purified attributes back to the visual space, thereby iteratively minimizing cross-modal discrepancies and mitigating overfitting under ambiguous labels. Such bidirectional alignment is particularly critical for ambiguous labels, as the mutual reinforcement between modalities effectively prevents overfitting to erroneous annotations.

\subsubsection{Attribute Mutual Information Representation Learning}

Owing to label ambiguity, we only observe a set of candidate semantic vectors rather than the true class semantics. This injects noise into the semantics, making it crucial to obtain discriminative attribute representations. Therefore, we introduce an MI-based contrastive optimization that preserves fine-grained attribute relationships in the embedding space.

Attribute feature selection is based on the entropy distribution of different classes of attributes. For each dataset, we calculate the Shannon entropy of each attribute column, which quantifies the contribution of each attribute to classification,
\begin{equation}
    H_k = -\sum_{c=1}^{Q} p(s_k^c) \log p(s_k^c),
\end{equation}
where $p(s_k^c)$ is the normalized occurrence probability. An adaptive threshold $\mu=\text{median}({H_1, ..., H_K})$ selects the top 50\% most discriminative attributes$\left(\mathcal{A}=\left\{\hat{\mathbf{a}}_k \mid H_{k}<\mu\right\}\right)$. For each selected attribute $\hat{\mathbf{a}}_k\in\mathcal{A}$, features from the same attribute form positives $\mathcal{A}_{+}$, and all other attribute features serve as negatives $\mathcal{A}_{-}$.

To mitigate the high-variance instability typically observed in MI estimation via variational $f$-divergences, we adopt a stabilized training strategy based on the \emph{Jensen--Shannon} (JS) divergence \cite{poole2019variational}. Specifically, we train the critic $V(\cdot)$ using a JS-based objective, which provides more stable gradients compared to directly optimizing exponential-form bounds.
During the forward pass, the MI value is computed using an NWJ-style estimator,
\begin{equation}
I_{MI} = \mathbb{E}_{\mathbb{P}}[V(\mathcal{A}, \mathcal{A}_{+})] - \mathbb{E}_{\mathbb{Q}}[\exp(V(\mathcal{A}, \mathcal{A}_{-}))] + 1,
\end{equation}
We encourage high MI between samples and their positives and low MI with negatives, and define the attribute MI loss as
\begin{equation} 
    \mathcal{L}_{AMI} = I_{{MI}},
\end{equation}

\begin{table*}
\centering
\caption{The comparison result (\%) with other ZSL methods on benchmarks with different degrees of random noise. The \textbf{best} and \underline{second-best} results are highlighted, respectively.}
\label{tab:pzsl}
\resizebox{\textwidth}{!}{%
  \begin{tabular}{*{18}{c}}
    \toprule
    \multirow{3}{*}{Datasets} & \multirow{3}{*}{Methods} & \multicolumn{4}{c}{$q=0.01$} & \multicolumn{4}{c}{$q=0.03$} & \multicolumn{4}{c}{$q=0.05$} & \multicolumn{4}{c}{$q=0.07$} \\
    \cmidrule(lr){3-6} \cmidrule(lr){7-10} \cmidrule(lr){11-14} \cmidrule(lr){15-18}
    & & \multicolumn{1}{c}{CZSL} & \multicolumn{3}{c}{GZSL} & \multicolumn{1}{c}{CZSL} & \multicolumn{3}{c}{GZSL} & \multicolumn{1}{c}{CZSL} & \multicolumn{3}{c}{GZSL} & \multicolumn{1}{c}{CZSL} & \multicolumn{3}{c}{GZSL} \\
    \cmidrule(lr){3-3} \cmidrule(lr){4-6} \cmidrule(lr){7-7} \cmidrule(lr){8-10} \cmidrule(lr){11-11} \cmidrule(lr){12-14} \cmidrule(lr){15-15} \cmidrule(lr){16-18}
    & & $T1$ & $U$ & $S$ & $H$ & $T1$ & $U$ & $S$ & $H$ & $T1$ & $U$ & $S$ & $H$ & $T1$ & $U$ & $S$ & $H$ \\
    \midrule
    \multirow{5}{*}{AwA2} & GEM-ZSL \cite{liu2021goal} & 33.8 & 33.2 & 43.0 & 37.5 & 28.6 & 25.8 & 32.0 & 28.6 & 21.8 & 21.5 & 34.0 & 26.3 & 18.5 & 18.2 & 27.8 & 22.0 \\
    & Transzero \cite{chen2022transzero} & 65.4 & 61.2 & \underline{76.5} & \underline{68.0} & \underline{65.5} & \underline{62.6} & 68.8 & \underline{65.6} & \underline{63.4} & \underline{61.4} & 55.2 & 58.1 & \underline{61.2} & \underline{55.8} & 41.9 & 47.9 \\
    & MSDN \cite{chen2022msdn} & \underline{66.4} & 59.3 & 75.1 & 66.3 & 62.2 & 54.6 & \underline{82.1} & \underline{65.6} & 57.9 & 52.8 & 74.4 & 61.8 & 57.0 & 45.0 &  \textbf{84.4} & 58.7 \\
    & CoAR-ZSL \cite{du2023boosting} & 62.2 & \underline{61.4} & 70.6 & 65.7 & 63.2 & 62.1 & 68.2 & 65.0 & 60.6 & 59.6 & 65.6 & \underline{62.5} & 53.1 & 52.1 & 66.8 & 58.5 \\
    & CvSC \cite{chen2024causal} & 52.7 & 49.0 & \textbf{85.1} & 62.2 & 50.8 & 49.2 & \textbf{83.2} & 61.8 & 50.0 & 48.5 & \textbf{81.2} & 60.7 & 48.8 & 47.8 & 77.6 & \underline{59.2} \\
    & \textbf{DVSA} & \textbf{76.4} & \textbf{75.1} & 74.1 & \textbf{74.6} & \textbf{73.7} & \textbf{75.5} & 73.2 & \textbf{74.4} & \textbf{73.6}& \textbf{73.0} & \underline{75.0} & \textbf{74.0} & \textbf{70.4} & \textbf{68.4} & \underline{80.3} & \textbf{73.9} \\
    \midrule
    \multirow{5}{*}{CUB} & GEM-ZSL \cite{liu2021goal} & 56.9 & 49.2 & 49.7 & 49.4 & 39.9 & 34.9 & 36.9 & 35.9 & 34.8 & 27.3 & 28.1 & 27.7 & 23.0 & 19.0 & 20.2 & 19.6 \\
    & Transzero \cite{chen2022transzero} & 62.9 & 51.0 & 56.0 & 53.4 & 39.4 & 25.9 & 38.0 & 30.8 & 31.1 & 21.5 & 33.0 & 26.0 & 28.7 & 14.8 & \underline{29.7} & 19.8 \\
    & MSDN \cite{chen2022msdn} & 56.5 & 47.1 & 59.1 & 52.4 & 39.1 & 27.9 & 49.8 & 35.7 & 27.6 & 21.4 & \underline{45.0} & 29.0 & 24.1 & 19.1 & 29.0 & \underline{23.0} \\
    & CoAR-ZSL \cite{du2023boosting} & 60.0 & 50.6 & 51.1 & 50.8 & 42.4 & 36.9 & 40.8 & 38.8 & 37.5 & 27.1 & 30.2 & 28.6 & 30.9 & 20.6 & 22.3 & 21.4 \\
    & CvSC \cite{chen2024causal} & \underline{67.9} & \underline{55.1} & \underline{68.8} & \underline{61.2} & \underline{58.3} & \underline{53.1} & \underline{50.2} & \underline{51.6} & \underline{50.0} & \underline{44.7} & 30.1 & \underline{36.0} & \underline{36.6} & \underline{36.5} & 15.2 & 21.5 \\
    & \textbf{DVSA} & \textbf{71.9} & \textbf{65.8} & \textbf{72.5} & \textbf{69.0} & \textbf{69.4}& \textbf{61.1} & \textbf{74.4} & \textbf{67.1} & \textbf{70.7} & \textbf{62.5} & \textbf{71.4} & \textbf{66.7} & \textbf{66.1} & \textbf{58.0} & \textbf{67.3} & \textbf{62.3} \\
    \midrule
    \multicolumn{1}{c}{Datasets} & \multicolumn{1}{c}{Methods} & \multicolumn{4}{c}{$r=1$} & \multicolumn{4}{c}{$r=2$} & \multicolumn{4}{c}{$r=3$} & \multicolumn{4}{c}{$r=4$} \\
    \cmidrule(lr){1-2} \cmidrule(lr){3-6} \cmidrule(lr){7-10} \cmidrule(lr){11-14} \cmidrule(lr){15-18}
    \multirow{5}{*}{SUN} & GEM-ZSL \cite{liu2021goal} & 61.4 & 38.4 & \underline{35.3} & 36.8 & 58.0 & 37.4 & 29.3 & 31.9 & 57.0 & 31.8 & 29.3 & 30.5 & 56.2 & 30.2 & \underline{28.3} & 29.2 \\
    & Transzero \cite{chen2022transzero} & 58.5 & \underline{47.4} & 22.6 & 30.6 & 55.7 & 48.7 & 15.7 & 23.7 & 53.6 & \underline{48.5} & 11.0 & 17.9 & 53.9 & \underline{46.9} & 7.1 & 12.4 \\
    & MSDN \cite{chen2022msdn} & 61.1 & 50.1 & 21.0 & 29.6 & 59.5 & \underline{48.9} & 16.4 & 24.6 & 57.3 & 45.8 & 11.9 & 18.8 & 54.9 & 35.5 & 10.3 & 16.0 \\
    & CoAR-ZSL \cite{du2023boosting} & 60.9 & 43.5 & 34.3 & 38.4 & 61.7 & 42.6 & 30.1 & 35.3 & 57.6 & 37.7 & \underline{29.8} & 33.3 & 54.9 & 35.4 & 27.3 & \underline{30.8} \\
    & CvSC \cite{chen2024causal}  & \textbf{68.2} & \textbf{64.8} & 32.8 & \textbf{43.6} & \textbf{67.4} & \textbf{64.3} & \underline{30.2} & \textbf{41.1} & \textbf{66.6} & \textbf{65.3} & 24.0 & \underline{35.1} & \textbf{66.2} & \textbf{64.2} & 19.0 & 29.3\\
    & \textbf{DVSA} & \underline{62.1} & 44.4 & \textbf{36.6} & \underline{40.1} & \underline{63.3} & 42.5 & \textbf{35.4} & \underline{38.6} & \underline{60.9} & 38.1 & \textbf{35.8} & \textbf{36.9} & \underline{60.7} & 35.6 & \textbf{32.7} & \textbf{34.1} \\
    \bottomrule
  \end{tabular}%
}
\end{table*}

\subsection{Dynamic Label Disambiguation Mechanism}

The label disambiguation mechanism aims to resolve ambiguity in the candidate label set by identifying the most probable ground-truth label for each instance. The updated labels are then used to guide the learning process of the model, gradually narrowing the semantic gap between instances and candidate labels. This mechanism is particularly crucial in scenarios with ambiguous labels, as it enables the model to iteratively refine its supervision signal, thereby enhancing its generalization capability.

For input image $\mathbf{x}_i$, the global visual features $\mathbf{v}$ is obtained via Global Average Pooling (GAP) on backbone features $\mathbf{f}_i$. The class prototypes $\mathbf{p}$ are obtained by embedding the class semantic vector $\mathbf{S}$ into the visual space. Then the visual prediction probability matrix $\mathbf{M}=[m_{ij}]_{N \times Q}$ is denoted by,
\begin{equation} 
    m_{ij}=\frac{\exp \left(\tau \cdot \cos \left(\mathbf{v}_i, \mathbf{p}_j\right)\right)}{\sum_{k=1}^{Q} \exp \left(\tau \cdot \cos \left(\mathbf{v}_i, \mathbf{p}_j\right)\right)},
\end{equation} 
where $\tau$ is the scaling factor. Instead of relying on pre-determined values, we update soft labels via dynamic disambiguation in the process of iterative learning,
\begin{gather} 
    \tilde{l}_{ij}^{t+1}=\left\{\begin{array}{cl}
    \mathbf{U}_{ij}^{t} / \sum_{y_{c} \in L_{i}} \mathbf{U}_{ic}^{t} & \text { if } y_{j} \in L_{i} \\0 & \text { otherwise }\end{array}\right.,
\end{gather} 
where $\mathbf{U}_{ij}^{t}=(1-\alpha)\mathbf{U}_{ij}^{t-1}+\alpha\mathbf{M}_{ij}^{t}$, with $\mathbf{U}_{ij}^{0}$ is initialized from the candidate labels. We set $\alpha=0.5$ to balance stable label refinement with responsiveness to new predictions. Thus the corrected labels for the $t$-th epoch are computed as a weighted combination of corrected labels from the previous epoch and current predictions. The corrected labels are subsequently passed into the model for training as the updated supervision signal in the next iteration. To further mitigate label ambiguity, we introduce a correction factor that measures the alignment between refined visual features $\hat{\mathbf{v}}_i$ and class prototypes $\mathbf{p}_{j}$,
\begin{equation}
		\omega _{ij}^{t} = \text{Softmax}\left(\frac{\hat{\mathbf{v}}_i^{t} \mathbf{p}_{j}^{T}}{\left\|\hat{\mathbf{v}}_i^{t}\right\|_{2}\left\|\mathbf{p}_{j}^{T}\right\|_{2}}\right),
\end{equation} 
This term acts as a weight in the loss,
\begin{equation}
		\mathcal{L}_{vis}=-\sum_{i=1}^{N}\sum_{j=1}^{Q} \left((1+\omega _{ij}^{t}) \tilde{l}_{ij}^{t}\log {m}_{ij}^{t}\right),
\end{equation} 

In parallel, we compute semantic confidence from refined attributes $\hat{\mathbf{a}}_i$ and class semantic embeddings $\mathbf{s}_j$,
\begin{equation} 
    m^{'}_{ij}=\frac{\exp \left(\tau \cdot \cos \left(\hat{\mathbf{a}}_i, \mathbf{s}_j\right)\right)}{\sum_{k=1}^{Q} \exp \left(\tau \cdot \cos \left(\hat{\mathbf{a}}_i, \mathbf{s}_j\right)\right)},
\end{equation} 
leading to a semantic cross-entropy loss
\begin{equation} 
		\mathcal{L}_{sem}=-\sum_{i=1}^{N}\sum_{j=1}^{Q} (\tilde{l}_{ij}^{t}\log {m}_{ij}^{'t}).
\end{equation} 

Adaptive cross-entropy loss combines both modalities with equal importance, i.e., $\mathcal{L}_{ACE}=\mathcal{L}_{vis}+\mathcal{L}_{sem}$. This mechanism ensures that the label information is progressively refined, mitigating the impact of noise during training.

\subsection{Model Optimization and Inference}

We optimize DVSA by minimizing a joint objective of adaptive cross-entropy and attribute mutual information loss,
\begin{equation} 
	\begin{split}
    \mathcal{L}=\mathcal{L}_{ACE}-\mathcal{L}_{AMI}.
	\end{split}
\end{equation} 
After training the model, we perform nearest-prototype classification in a cosine similarity space:

\begin{equation}
    y^{*} = \arg\max_{y_{c} \in \mathcal{Y}^{t}}
    \big(\mathbf{v} \cdot \mathbf{p}_{c} - \gamma \,\mathbb{I}[y_{c} \in \mathcal{Y}^{s}]\big).
\end{equation}
where $\mathbb{I}(\cdot)$ denotes an indicator function. For CZSL, we classify only over unseen labels by setting $\mathcal{Y}^{t}=\mathcal{Y}^{u}$ and $\gamma=0$. For GZSL, test instances may come from both seen and unseen classes, i.e., $\mathcal{Y}^{t}=\mathcal{Y}^{s}\cup\mathcal{Y}^{u}$, so we adopt calibrated stacking with $\gamma>0$ to reduce the bias toward seen classes.

\begin{table}[t]
  \centering
  \caption{The comparison result (\%) with other ZSL methods on benchmarks. The \textbf{best} and \underline{second-best} results are highlighted, respectively.}
  \label{tab:zsl}
  \resizebox{.99\columnwidth}{!}{
  \begin{tabular}{cccccccc}
    \toprule
    \multirow{2}{*}{Methods} & \multirow{2}{*}{Venue} & \multicolumn{2}{c}{AwA2} & \multicolumn{2}{c}{CUB} & \multicolumn{2}{c}{SUN} \\
    \cmidrule(lr){3-4} \cmidrule(lr){5-6} \cmidrule(lr){7-8}
    & & $T1$ & $H$ & $T1$ & $H$ & $T1$ & $H$ \\
    \midrule
    \multicolumn{8}{c}{Generative-based Methods} \\
    \cmidrule(lr){1-8}
    HSVA \cite{chen2021hsva} & NeurIPS'21 & - & 66.8 & 62.8 & 55.3 & 63.8 & 43.3 \\
    CE-GZSL \cite{han2021contrastive} & CVPR'21 & 70.4 & 70.0 & 77.5 & 65.3 & 63.3 & 43.1 \\
    ICCE \cite{kong2022compactness} & CVPR'22 & 72.7 & 72.8 & \underline{78.4} & 66.4 & - & - \\
    BSeGN \cite{xie2022leveraging} & TNNLS'22 & 71.5 & 67.4 & 65.3 & 58.0 & \underline{66.4} & 42.9 \\
    AREES \cite{liu2022zero} & TNNLS'22 & 73.6 & 66.1 & 65.7 & 55.2 & 64.3 & 42.2 \\
    DGCNet \cite{zhang2023dual} & TCSVT'23 & 74.1 & 62.1 & 71.6 & 53.2 & 62.6 & 30.8 \\
    D$^3$GZSL \cite{wang2024data} & AAAI'24 & - & 70.1 & - & 67.8 & - & - \\
    VFD-ZSL \cite{he2024visual} & ICME'24 & 73.6 & 68.1 & 62.2 & 56.9 & 63.8 & \underline{43.5} \\
    \midrule
    \multicolumn{8}{c}{Embedding-based Methods} \\
    \cmidrule(lr){1-8}
    APN \cite{xu2020attribute} & NeurIPS'20 & 68.4 & 65.5 & 72.0 & 67.2 & 61.6 & 37.6 \\
    GEM-ZSL \cite{liu2021goal} & CVPR'21 & 67.3 & 70.6 & 77.8 & 70.4 & 62.8 & 36.9 \\
    A-RSR \cite{liu2022rethink} & TNNLS'22 & 69.0 & 65.6  & 72.1 & 68.0 & 64.0 & 39.0 \\
    Transzero \cite{chen2022transzero} & AAAI'22 & 70.1 & 70.2 & 76.8 & 68.8 & 65.6 & 40.8 \\
    MSDN \cite{chen2022msdn} & CVPR'22 & 70.1 & 67.7 & 76.1 & 68.1 & 65.8 & 41.3 \\
    CoAR-ZSL \cite{du2023boosting} & TNNLS'23 & \underline{74.1} & \underline{73.2} & \textbf{79.2} & \textbf{74.0} & \textbf{66.7} & 43.4 \\
    BGSNet \cite{li2023diversity}& TMM'23 & 69.1 & 69.9 & 73.3 & 66.7 & 63.9 & 39.0 \\
    ZS-VAT \cite{han2024zs} & TNNLS'25 & 62.8 & 70.9 & 66.4 & 69.4 & 42.9 & 40.7 \\
    \cmidrule(lr){1-8}
    DVSA & Ours & \textbf{76.7} & \textbf{75.8} & 74.9 & \underline{70.8} & 64.3 & \textbf{44.8} \\
  \bottomrule
  \end{tabular}
}
\vskip -0.1in
\end{table}

\section{Experiments}

\subsection{Experimental Setup}

\paragraph{Datasets}
We evaluate DVSA on three widely used datasets, i.e., \textbf{CUB} \cite{welinder2010caltech}, \textbf{SUN} \cite{patterson2014sun}, and \textbf{AwA2} \cite{xian2018zero}. The seen-unseen class division is set according to \cite{xian2018zero}.
Following the synthesis method of PLL datasets \cite{zhang2015solving}, we construct candidate label sets by adding noisy labels to each instance. For CUB and AwA2, we use $Q-1$ independent decisions with $q \in \left\{  0.01,0.03,0.05,0.07 \right\}$ controlling the inclusion probability of each noisy label. For SUN dataset, $r \in {1,2,3,4}$ specifies the number of false positive labels attached to each ambiguously labeled instance. Larger $q$ or $r$ yields more noisy labels and thus a harder disambiguation problem. On all three benchmarks, irrelevant labels are randomly sampled to form the candidate sets together with the ground-truth label. 

\paragraph{Metrics}
For CZSL, we measure performance using class-wise Top-$1$ accuracy on unseen classes. For GZSL, we employ the harmonic mean $H=(2 \times S \times U) /(S+U)$ to evaluate the performance, where $S$ and $U$ denote the Top-$1$ accuracy of seen and unseen classes, respectively.

\begin{table}[t]
  \centering
  \caption{Ablation study of different components on CUB and AwA2 datasets with $q=0.05$. ('/' indicates that only regular cross entropy training is used, without label updates)}
  \resizebox{.99\columnwidth}{!}{
  \begin{tabular}{ccccccccccccc}
    \toprule
    \multirow{2}{*}{VTA} & \multirow{2}{*}{$\mathcal{L}_{vis}$} & \multirow{2}{*}{$\mathcal{L}_{sem}$} & \multirow{2}{*}{$\omega $} & \multirow{2}{*}{$\mathcal{L}_{AMI}$} 
    & \multicolumn{4}{c}{CUB} 
    & \multicolumn{4}{c}{AwA2} \\
    \cmidrule(lr){6-9} \cmidrule(lr){10-13}
    & & & & & $T1$ & $U$ & $S$ & $H$ & $T1$ & $U$ & $S$ & $H$ \\
    \midrule
    $\times$ & / & $\times$ & $\times$ & $\times$ & 25.0 & 17.7 & 22.7 & 19.9 & 42.8 & 42.0 & 62.4 & 50.2 \\
    $\checkmark$ & / & $\times$ & $\times$ & $\times$ & 38.8 & 28.6 & 35.0 & 31.5 & 59.4 & 58.4 & 67.5 & 62.6 \\
    $\checkmark$ & $\checkmark$ & $\times$ & $\times$ & $\times$ & 51.0 & 45.9 & 51.2 & 48.4 & 60.4 & 59.0 & 71.3 & 64.6 \\
    $\checkmark$ & $\checkmark$ & $\checkmark$ & $\times$ & $\times$ & 54.2 & 47.6 & 51.4 & 49.4 & 65.8 & 65.1 & 73.2 & 68.9 \\
    $\checkmark$ & $\checkmark$ & $\checkmark$ & $\times$ & $\checkmark$ & 62.9 & 57.2 & 67.0 & 61.7 & 70.8 & 68.7 & 71.0 & 69.8 \\
    $\checkmark$ & $\checkmark$ & $\checkmark$ & $\checkmark$ & $\times$ & 63.4 & 56.7 & 69.1 & 62.3 & 67.3 & 66.9 & 71.4 & 69.1 \\
    $\checkmark$ & $\checkmark$ & $\checkmark$ & $\checkmark$ & $\checkmark$ & \textbf{70.7} & \textbf{62.5} & \textbf{71.4} & \textbf{66.7} & \textbf{73.6} & \textbf{73.0} & \textbf{75.0} & \textbf{74.0} \\
\bottomrule
\end{tabular}
}
\label{tab:abl}
\end{table}

\subsection{Comparison with State-of-the-Arts}
We first compare DVSA with strong embedding-based ZSL baselines under different levels of label ambiguity. As reported in Tab.~\ref{tab:pzsl}, DVSA achieves the best or second-best performance on all three datasets and across all ambiguity level in both CZSL and GZSL metrics. As the ambiguity level increases, all methods suffer performance degradation, but the drop of DVSA is substantially smaller than that of these. This indicates that the proposed modules jointly provide strong robustness to ambiguous labels.

To further assess the effectiveness of DVSA in the standard ZSL scenario without synthetic ambiguity, we compare it with recent generative-based and embedding-based methods on the original benchmarks. As shown in Tab.~\ref{tab:zsl}, DVSA achieving the best or near-best $T1$ and $H$ on CUB, AwA2, and SUN. These results demonstrate that DVSA not only excels in challenging ambiguous-label settings, but also remains competitive with state-of-the-art ZSL models on clean benchmarks, confirming the general effectiveness of the proposed framework.

\subsection{Ablation Study}

We conduct ablations on fine-grained CUB and coarse-grained AwA2 datasets with $q=0.05$, as summarized in Tab.~\ref{tab:abl}. Starting from a naive baseline that directly optimizes standard cross-entropy on candidate labels without bidirectional attention, dynamic disambiguation, semantic branch, calibration factor $\omega$, or MI loss, the model exhibits limited performance, showing that naive learning from ambiguous labels is suboptimal. Equipping this baseline with the VTA block already brings a noticeable gain. Introducing the visual adaptive cross-entropy loss with dynamically updated soft labels on top of VTA then yields a clear improvement across all metrics, indicating that iterative label refinement provides much cleaner supervision. Adding the semantic pathway further improves results by enforcing consistency between attribute-based and visual predictions. Comparing variants with and without $\omega$ and $\mathcal{L}_{AMI}$ shows that both calibration and MI-based contrastive optimization contribute additional gains. The calibration factor $\omega$ links the loss reweighting to the ATV block, while $\mathcal{L}_{AMI}$ exerts a stronger influence by producing more discriminative attribute embeddings. Combining all components yields the full DVSA, which achieves the best results on both datasets, confirming that these modules are highly complementary.

\begin{figure}[t]
  \centering
  \includegraphics[width=.85\columnwidth]{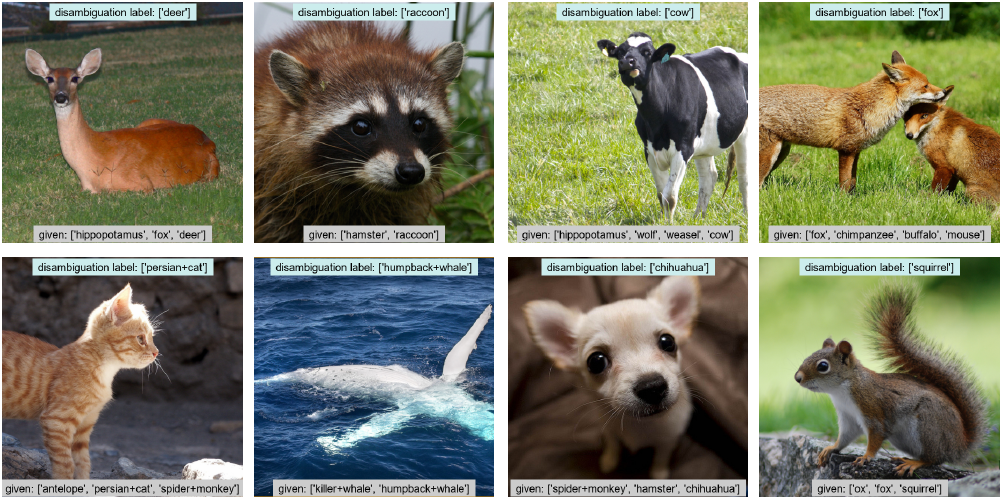}
  \caption{Visualization of disambiguation results of DVSA on AwA2 dataset.}
  \label{fig:disambiguation}
\end{figure}

\begin{figure}[t]
  \centering
  \includegraphics[width=.85\columnwidth]{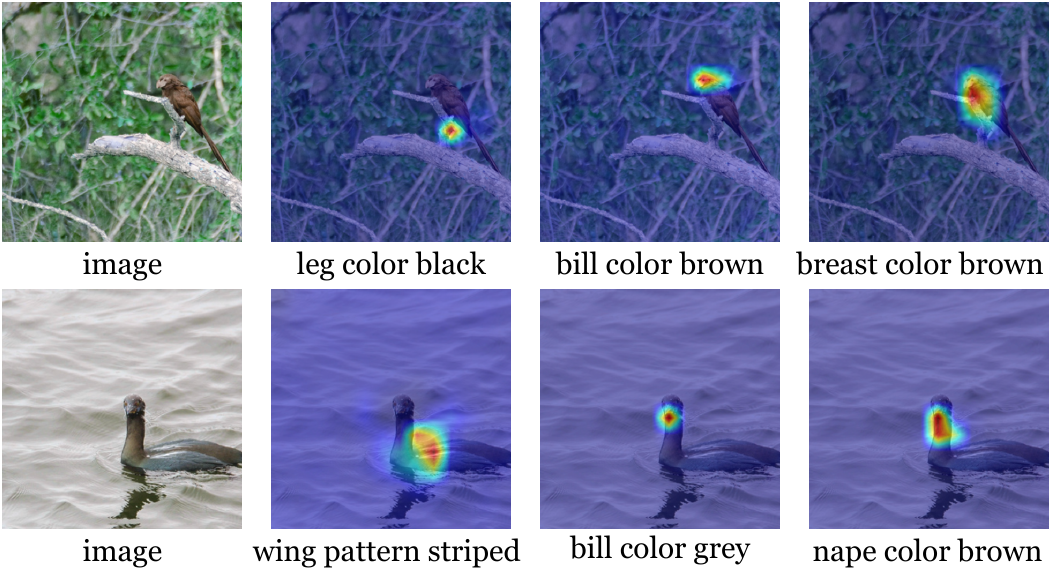}
  \caption{Visualization of attribute attention maps on CUB dataset.}
  \label{fig:attention}
  \vskip -0.1in
\end{figure}

\subsection{Visualization Analysis}

\paragraph{Disambiguation Process}
To elucidate the disambiguation mechanism, we conducted disambiguation visualization for the seen class on AwA2 dataset with $q=0.07$. As illustrated in Fig.~\ref{fig:disambiguation}, the candidate label sets are displayed below each image instance, while the corresponding disambiguation results are shown above. The results clearly demonstrate that DVSA maintains robust disambiguation accuracy despite substantial label ambiguity. 

\paragraph{Attribute Localization}
Fig.~\ref{fig:attention} qualitatively evaluates the VTA attention block through visualized attention maps. The results demonstrate that our method localizes attribute-relevant regions across diverse semantic concepts with high precision.

\section{Conclusion}
In this paper, we presented DVSA, a novel framework that addresses ambiguous labels in zero-shot learning through dynamic visual-semantic alignment. By integrating bidirectional visual-semantic attention and mutual information optimization, DVSA enhances the alignment between visual features and semantic attributes, improving the robustness of the model to label noise. Additionally, the dynamic label disambiguation mechanism progressively refines supervision signals, ensuring accurate predictions even in the presence of ambiguous annotations. Comprehensive experiments validate that DVSA achieves superior performance compared to existing methods, particularly in high-noise scenarios.


\bibliographystyle{IEEEbib}
\bibliography{icme2026references}

\end{document}